# Narrative-Guided Reinforcement Learning: A Platform for Studying Language Model Influence on Decision Making

**Anup Tuladhar**
Department of Pediatrics
University of Calgary
anup.tuladhar@ucalgary.ca

**Araz Minhas**
Department of Pediatrics
University of Calgary
araz.minhas@ucalgary.ca

**Adam Kirton**
Department of Pediatrics
University of Calgary
adam.kirton@ahs.ca

**Eli Kinney-Lang**
Department of Biomedical Engineering
University of Calgary
eli.kinneylang@ucalgary.ca

## Abstract

We present a preliminary experimental platform that explores how narrative elements might shape AI decision-making by combining reinforcement learning (RL) with language model reasoning. While AI systems can now both make decisions and engage in narrative reasoning, these capabilities have mostly been studied separately. Our platform attempts to bridge this gap using a dual-system architecture to examine how narrative frameworks could influence reward-based learning. The system comprises a reinforcement learning policy that suggests actions based on past experience, and a language model that processes these suggestions through different narrative frameworks to guide decisions. This setup enables initial experimentation with narrative elements while maintaining consistent environment and reward structures. We implement this architecture in a configurable gridworld environment, where agents receive both policy suggestions and information about their surroundings. The platform's modular design facilitates controlled testing of environmental complexity, narrative parameters, and the interaction between reinforcement learning and narrative-based decisions. Our logging system captures basic decision metrics, from RL policy values to language model reasoning to action selection patterns. While preliminary, this implementation provides a foundation for studying how different narrative frameworks might affect reward-based decisions and exploring potential interactions between optimization-based learning and symbolic reasoning in AI systems.

# 1   Introduction

Narratives fundamentally shape how humans think and make decisions. We naturally view our experiences, goals, and choices through narrative lenses that do more than just describe our decisions – they actively shape how we see options, evaluate consequences, and understand our choices [1]. These narrative frameworks provide crucial context that guides both our decision-making and how we later explain our choices to ourselves and others. Given the central role of narrative in human cognition and decision-making, it likely has important implications for developing AI systems that aim to replicate or enhance human-like reasoning capabilities – implications that remain largely unexplored [2].

Recent advances in artificial intelligence have made progress in two key areas relevant to decision-making: reinforcement learning and language modeling. Reinforcement learning has proven effective at developing agents that can master complex tasks through trial-and-error learning [3]. Meanwhile, language models have advanced in their ability to generate sophisticated narratives and demonstrate reasoning capabilities [4,5]. However, these two powerful approaches have largely developed independently, with limited exploration of how they might complement each other. While each approach has its strengths – RL's ability to learn from experience [6] and language models' capacity for contextual reasoning [7] – their potential synergy in decision-making tasks remains underexplored.

To begin exploring this gap, we've developed a preliminary experimental platform that combines reinforcement learning and language modeling in a dual-system architecture [8,9]. The platform attempts to augment traditional RL-based navigation with language model processing that can interpret and potentially override policy suggestions based on narrative context and environmental observations. This approach explores how systems might consider actions not just in terms of learned rewards, but through the lens of character-driven narratives. Our initial contribution is the development of a platform that may enable systematic study of how different narrative frameworks – from simple instruction-following to rich character-based reasoning – could influence decision-making while maintaining consistent environmental parameters and reward structures.

We present initial experiments with the platform comparing baseline RL performance against various narrative-guided approaches across different environmental complexities. These preliminary tests explore how narrative frameworks might affect both task performance and behavioral patterns, particularly in challenging scenarios where pure RL-based learning may be limited by training time or environmental complexity. The platform represents three potential steps forward: it demonstrates the feasibility of combining RL with narrative reasoning, provides a starting point for systematically studying how narratives might influence decisions, and offers a foundation for future empirical research into how narrative context could affect decision-making in complex environments.

# 2   Methods

Our platform combines reinforcement learning with language model-based narrative reasoning in a dual-system architecture. The RL module generates action suggestions based on its interactions with the environment, while the language model processes these suggestions through narrative prompts that frame each decision within a story context. We developed a logging system to capture basic metrics including success rates, step counts, and narrative processing outputs to enable initial analysis of how different narrative frameworks might affect decision-making patterns.

The system follows a basic interaction loop: at each decision point, the RL agent suggests an action based on its current Q-learning policy. This suggested action, along with observations about adjacent grid cells, passes to the language model. The language model processes this information through a configurable narrative framework, which can either reinforce or override the RL agent's suggestion based on its narrative interpretation. This design represents an initial attempt to explore how narrative context might influence the interpretation of reward-based suggestions.

For this preliminary study, we implemented our system in a simple gridworld environment, chosen primarily for its controllable parameters and straightforward success metrics. We tested the system across square grids ranging from 5x5 to 11x11 cells, with 30-40% of cells randomly designated as obstacles. Agents attempt to navigate from a start position to a goal position while avoiding obstacles. The RL component uses basic Q-learning with randomly initialized parameters [10], focusing on short training periods (10 episodes) to explore potential narrative influence rather than optimal policy learning. For narrative processing, we conducted initial tests with multiple language models, using GPT-4o-mini for our primary experiments due to its practical balance of performance and computational efficiency.

Our initial experimental design explored three aspects of potential narrative influence. First, we collected baseline data comparing pure RL agents against a basic LLM+RL implementation using direct navigation instructions. This baseline used minimal narrative framing, simply instructing the agent about its gridworld environment and the meaning of RL policy suggestions. Second, we tested several grid sizes and obstacle densities to begin identifying conditions where narrative guidance might show impact. Finally, we tested several narrative frameworks, including a Theseus-inspired labyrinth navigator, a Sherlock Holmes detective scenario, and a Westworld-inspired AI agent, while maintaining consistent environmental parameters across conditions.

To begin evaluating system behavior, we tracked basic performance metrics: success rate (percentage of episodes reaching the goal), average steps to completion (for successful episodes only), computational overhead of LLM processing, and adherence to or deviation from RL policy suggestions. While preliminary, this data collection approach allowed us to make initial observations about both task performance and behavioral patterns across different narrative conditions and environmental complexities.

## 3 Results

Our experimental evaluation revealed both baseline performance characteristics of the dual-system architecture and specific effects of narrative frameworks on agent behavior. Initial testing across different grid sizes established fundamental performance patterns of the LLM+RL integration. While the addition of LLM processing introduced significant computational overhead (5-30 minutes for 10 episodes compared to <1 second for RL-only), the LLM+RL agents demonstrated more efficient path-finding per interaction step. This efficiency gain was particularly evident in complex environments (7x7 grids and larger), where LLM+RL agents achieved in 10 episodes performance comparable to what RL-only agents reached after 100 episodes of training (Figure 1).

Having established these baseline characteristics, we focused on our primary research question: the influence of narrative frameworks on decision-making. We evaluated this influence through two key metrics: episode success rate (percentage of episodes where the agent reached the goal) and average steps to completion (mean number of steps taken during successful episodes). Testing focused on 7x7 grids with 30% or 40% obstacle density, as these parameters provided sufficient complexity to differentiate agent performance while remaining computationally tractable.

Direct comparison of narrative frameworks revealed systematic differences in both performance metrics. The baseline LLM+RL agent, operating under simple task instructions without narrative context, established reference performance levels (Figure 1C). Narrative-driven agents consistently matched or exceeded this baseline in terms of episode success rate, but showed marked differences in path efficiency (Figure 1D–F). The Theseus labyrinth navigator (Figure 1D) and Sherlock Holmes (Figure 1E) detective frameworks demonstrated comparable success rates to the baseline, but with reduced average steps to completion in successful episodes. Most notably, the Westworld-inspired AI agent framework consistently outperformed other narratives across both metrics, achieving both higher success rates and lower average steps to completion (Figure 1F).

Analysis of agent behavior logs revealed that narrative frameworks influenced not just performance metrics but also decision-making patterns. While all LLM+RL agents had access to the same environmental information and RL policy suggestions, their interpretation and use of this information varied systematically with narrative context. The Westworld AI framework, for example, demonstrated more consistent integration of RL suggestions with environmental observations, potentially explaining its superior performance metrics. These behavioral differences persisted across multiple environmental configurations, suggesting robust influence of narrative framing on decision-making strategies [11].

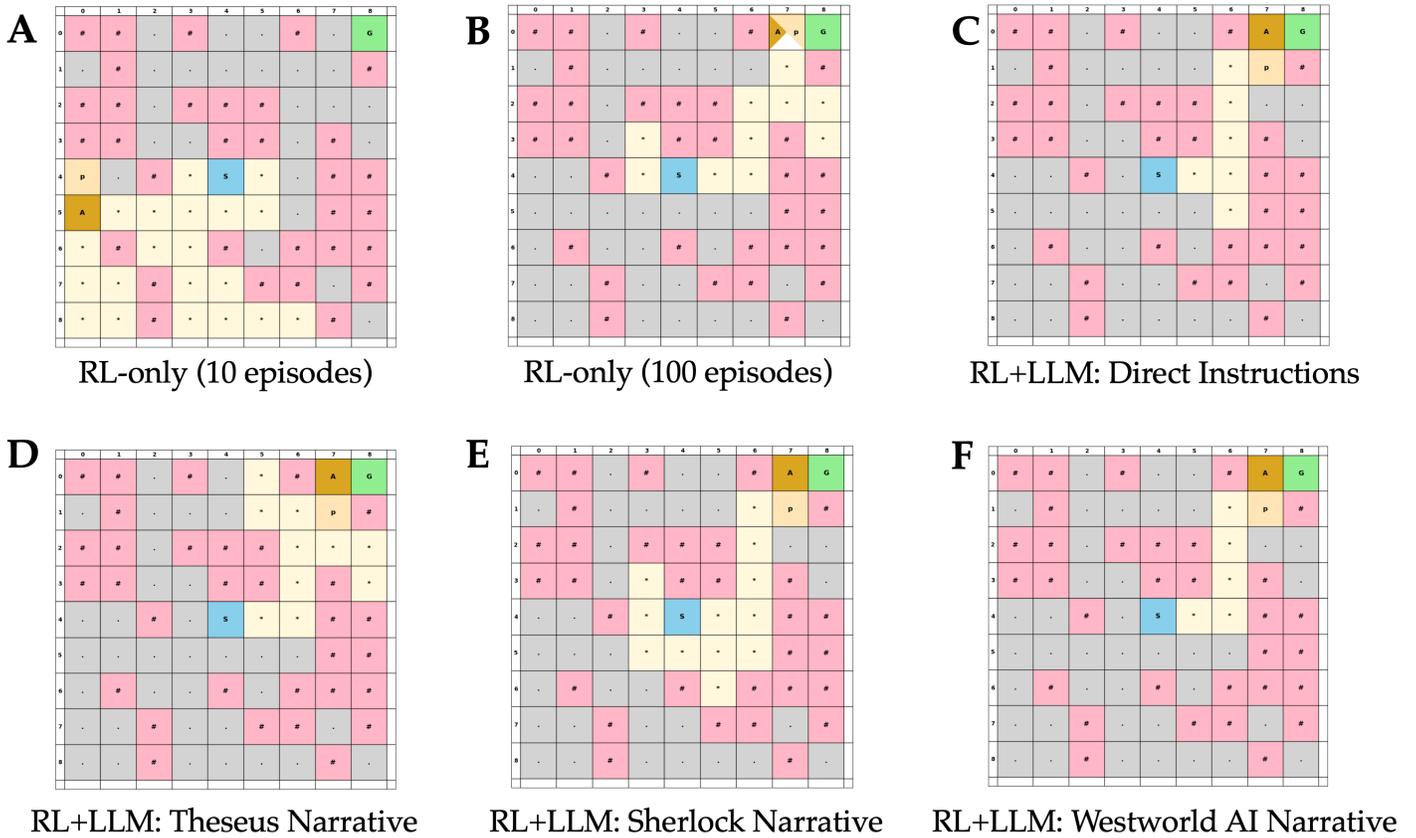

Figure 1: Example agents navigating gridworld environment after training. (A) RL-only policy after 10 episodes of training. (B) RL-only after 100 episodes of training. (C–F) RL+LLM policies after 10 episodes of training, with (C) Direct instructions of the task, (D) Theseus narrative inspired instructions, (E) Sherlock narrative inspired instructions, or (F) Westworld AI narrative inspired instructions

## 4  Discussion

Our preliminary experiments demonstrate the basic feasibility of combining narrative frameworks with reinforcement learning in a controlled experimental platform. While our initial results suggest some interesting patterns in how narrative context might influence agent behavior, these findings should be considered strictly exploratory. The observed differences between direct instruction and narrative-guided approaches, particularly in more complex environments, indicate this may be a promising direction for further research. However, substantially more rigorous testing is needed before drawing any conclusions about the impact of narrative on agent decision-making.

The platform itself represents a first step toward systematically studying narrative influence in AI decision-making systems [12]. Our implementation demonstrates that it is technically feasible to combine reinforcement learning with language model processing in a way that allows controlled experimentation. However, significant work remains to establish the robustness and reproducibility of our initial observations. Key questions remain about how different model scales affect performance, how consistently narrative frameworks influence behavior across different environmental conditions, and whether more sophisticated reasoning approaches might yield different results.

This work opens numerous avenues for future investigation. Important next steps include evaluating performance across a range of language model sizes, from smaller specialized models to larger general-purpose ones, to better understand the relationship between model capacity and narrative processing. Exploring more complex reasoning schemas, such as chain-of-thought prompting and multi-agent architectures, could provide insight into how different approaches to narrative processing affect decision-making [13,14]. Additionally, investigating more sophisticated narrative frameworks and their interaction with various reinforcement learning algorithms could help establish whether our preliminary observations hold across different experimental conditions. While our current implementation focuses on a simple gridworld environment, the underlying architecture could potentially extend to more complex domains – though such extensions would require careful validation and likely present new technical challenges.